\documentclass[conference]{IEEEtran}
\usepackage{graphicx} 
\usepackage{ifthen}
\usepackage{amssymb}
\usepackage{amsmath}
\newcommand{\set}[1]{\mathcal{#1}}
\newcommand{\squishlisttwo}{
 \begin{list}{$\bullet$}
  { \setlength{\itemsep}{0pt}
     \setlength{\parsep}{0pt}
    \setlength{\topsep}{0pt}
    \setlength{\partopsep}{0pt}
    \setlength{\leftmargin}{1em}
    \setlength{\labelwidth}{1.5em}
    \setlength{\labelsep}{0.5em} } }
\newcommand{\squishend}{
  \end{list}  }

\newtheorem{thm}{Theorem}
\ifCLASSINFOpdf
\else
\fi
\newcounter{sol} 
\begin{document}
%
\title{Decision-Theoretic Coordination and Control for Active Multi-Camera Surveillance in\\Uncertain, Partially Observable Environments}

\author{\IEEEauthorblockN{Prabhu Natarajan, Trong Nghia Hoang, Kian Hsiang Low, and Mohan Kankanhalli}
\IEEEauthorblockA{Department of Computer Science, National University of Singapore, Republic of Singapore\\
\{prabhu, nghiaht, lowkh, mohan\}@comp.nus.edu.sg}}
\maketitle
\begin{abstract}
A central problem of surveillance is to monitor multiple targets moving in a large-scale, obstacle-ridden environment with occlusions.
This paper presents a novel principled Partially Observable Markov Decision Process-based approach to coordinating and controlling a network of active cameras for tracking and observing multiple mobile targets at high resolution in such surveillance environments.
Our proposed approach is capable of (a) maintaining a belief over the targets' states (i.e., locations, directions, and velocities) to track them, even when they may not be observed directly by the cameras at all times, (b) coordinating the cameras' actions to simultaneously improve the belief over the targets' states and maximize the expected number of targets observed with a guaranteed resolution, and (c) exploiting the inherent structure of our surveillance problem to improve its scalability (i.e., linear time) in the number of targets to be observed.
Quantitative comparisons with state-of-the-art multi-camera coordination and control techniques show that our approach can achieve higher surveillance quality in real time. 
The practical feasibility of our approach is also demonstrated using real AXIS 214 PTZ cameras.
\end{abstract}


%
\section{Introduction}
\label{sec:int}
Monitoring, tracking, and observing multiple mobile targets in a large-scale, obstacle-ridden environment (e.g., airport terminals, railway stations, bus depots, shopping malls, etc.) is a central problem of surveillance. It is often necessary to acquire high-resolution videos/images of these targets.
Traditionally, such high-quality surveillance is achieved by placing a large number of static cameras to completely cover the large environment. This is impractical in terms of equipment, installation, and maintenance costs. Recent works  (\cite{Krahnstoever_2008,Natarajan_2012,Starzyk_2011}) have employed a heterogeneous network of wide-view static camera(s) to detect and track the targets within the environment at low resolution and some active pan/tilt/zoom (PTZ) cameras to be directed and focused on these targets to observe them at high resolution. 
Such surveillance systems 
face two serious practical limitations:
(a) The obstacles in the environment (e.g., physical structures like walls, pillars, and barriers) are likely to occlude the fields of view (fov) of the static cameras and hence they cannot detect or track the targets that reside within these \emph{occluded regions}. Since the surveillance system is not informed of these targets, the active cameras may not be directed to observe them; and (b) when the targets move further away from the low-resolution static cameras,
their measured locations become less accurate regardless of the calibration method. The vision algorithms to detect and recognize the targets also grow less reliable.

More importantly, the above limitations raise a practical implication affecting real-world multi-camera surveillance in general: The exact locations of the targets may not be observed directly by (or fully observable to) the cameras at all times. Such an environment is said to be \emph{partially observable} \cite{Kaelbling_1998}. Instead of introducing additional static cameras to resolve this issue of partial observability, we propose an alternative that maintains a probabilistic belief over the targets' possible locations, directions, and velocities in the environment. Our proposed alternative offers a practical advantage over (\cite{Krahnstoever_2008,Natarajan_2012,Starzyk_2011}) by eliminating the dependence on wide-fov static cameras to track the targets' locations 
and enabling the active PTZ cameras to perform dual roles of tracking the targets as well as observing them with high resolution.
Hence, we will focus on using (though not limited ourselves to) only active PTZ cameras in this paper.

This paper presents a novel principled decision-theoretic approach to coordinating and controlling a network of active cameras for tracking and observing multiple mobile targets at high resolution in uncertain, partially observable surveillance environments.
Our proposed approach stems from framing the surveillance problem formally using a rich class of decision making under uncertainty models called the \emph{Partially Observable Markov Decision Process} (POMDP) (Section~\ref{sec:sysarc}). Specifically, it overcomes the above limitations by (a) modeling a belief over the targets' states (i.e., locations, directions, and velocities) and updating the belief in a Bayesian paradigm (Section~\ref{sec:beltar}) based on
probabilistic models of the targets' motion (Section~\ref{sec:tramod}) and the active cameras' observations (Section~\ref{sec:obsmod}); (b) 
coordinating the active cameras' actions to simultaneously improve the belief over the targets' states and maximize the expected number of targets observed with a guaranteed pre-defined resolution (Sections~\ref{sec:objfun} and~\ref{sec:polcom});
and (c) exploiting the inherent structure of our surveillance problem to improve its scalability (i.e., linear time) in the number of targets to be observed (Section~\ref{sec:polcom}). Our POMDP-based approach is empirically evaluated in simulations over various realistic surveillance environments (Section~\ref{sec:simexp}) and with real AXIS $214$ PTZ cameras to demonstrate its practical feasibility (Section~\ref{sec:realexp}).

%
%

\section{Related Works}
\label{sec:relwor}
As mentioned earlier, existing multi-camera coordination and control techniques 
have to operate in a fully observable surveillance environment where the locations, directions, and velocities of all the targets can be directly observed/estimated by either using additional low-resolution static cameras and sensors (\cite{Costello_2005,Krahnstoever_2008,Natarajan_2012,Starzyk_2011,Ward_2009})  or configuring one or more active cameras to zoom out to their wide view (\cite{Huang_2011,Sommerlade_2010,Song_2008,Soto_2009}).
They use these targets' information to predict their trajectories in order to schedule, coordinate, and control the network of active cameras to focus on and observe these targets at high resolution. 

The major drawbacks of these techniques are: (a) They cannot be deployed in real-world surveillance environments with occlusions. In this case, they cannot observe the targets that reside in the occluded regions, hence limiting the active cameras' full surveillance capability. In contrast, our approach does not assume that all targets can be fully observed at every time instance, and hence models a belief of the targets' states to keep track of them when they are not observed by any of the cameras; (b) since the resolution of the wide-view static cameras is low, they often produce inaccurate locations of the targets. This in turn induces errors in targets' directions and velocities which consequently affect the prediction capability of existing surveillance systems.
On the other hand, our approach uses only active cameras to observe the targets at high resolution, thus allowing location errors to be kept minimal; and (c) many existing techniques have serious issues of scalability in the number of targets to be observed. 
Our approach extends our previous work \cite{Natarajan_2012} to achieve scalability  in partially observable surveillance environments.

\section{System Overview and Problem Formulation}
\label{sec:sysarc}
\begin{figure}[h]
 \centering
  \vspace{-2mm}
   \includegraphics[width=0.48\textwidth]{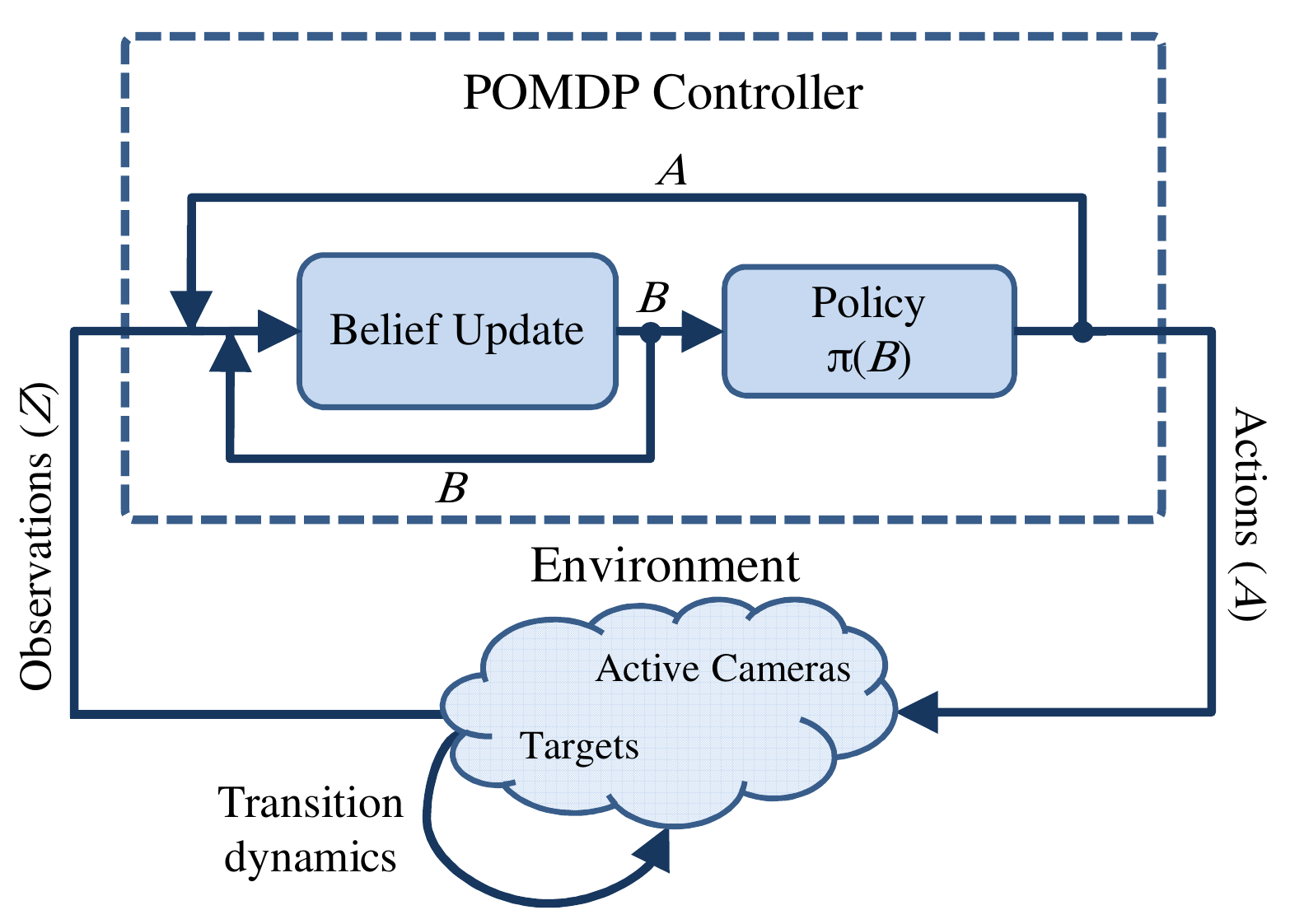}
  \vspace{-3mm}
  \caption{POMDP controller for coordinating active cameras to perform high-quality surveillance in a partially observable environment.}
  \label{fig:pomdpcont}
  \vspace{-2mm}
\end{figure}
A POMDP controller models the interaction between the active cameras and the partially observable surveillance environment. In particular, it is responsible for coordinating the cameras' actions to achieve a high-level surveillance goal which can be defined formally using a real-valued objective function and, in the context of this paper, is to maximize the number of targets observed with a guaranteed resolution (Section~\ref{sec:objfun}).
By calibrating the active cameras, they can determine the locations of the targets observed in their fov, which are communicated to the controller. The targets are assumed to be non-evasive and hence their motion cannot be controlled or influenced by the cameras.
The targets' correspondences across multiple cameras are resolved by distinct features like color and texture.

Formally, a POMDP controller is defined as a tuple ($\set{S}$, $\set{A}$, $\set{Z}$, $T_f$, $O_f$, $R$) consisting of
\squishlisttwo
\item a set $\set{S}$ of joint states of active cameras and targets in the surveillance environment (Section~\ref{sec:staact});
\item a set $\set{A}$ of joint actions of active cameras (Section~\ref{sec:staact});
\item a set $\set{Z}$ of joint observations of the targets taken by the cameras (Section~\ref{sec:staact}); 
\item a transition function $T_f : \set{S} \times \set{A} \times \set{S} \rightarrow [0,1]$ denoting the probability $P(S'|S,A)$ of going from the current joint state $S\in\set{S}$ to the next joint state $S'\in\set{S}$ using the joint action $A\in \set{A}$ (Section~\ref{sec:tramod});
\item an observation function $O_f : \set{S} \rightarrow [0,1]$ denoting the probability $P(Z|S)$ of observing the joint observation $Z\in\set{Z}$ given the joint state $S\in\set{S}$ (Section~\ref{sec:obsmod}); and 
\item a real-valued objective/reward function $R : S \rightarrow \mathbb{R}$ representing a high-level surveillance goal (Section~\ref{sec:objfun}). 
\squishend
At any given time, the exact state of the environment is not fully observable to the POMDP controller. Instead, it maintains a belief $B$ over the set $\set{S}$ of all possible states  (Section~\ref{sec:beltar}), that is, $B(S)$ is the probability that the environment is in the state $S\in\set{S}$ such that $\sum_{S\in\set{S}}{B(S)}=1$. As shown in Fig.~\ref{fig:pomdpcont}, at every time step, the POMDP controller issues an action $A\in\set{A}$ and makes an observation $Z\in\set{Z}$ from the environment. Based on the action $A$ and observation $Z$, the prior belief $B$ is updated by Bayes' rule to the posterior belief $B'$ as follows:
\vspace{-0mm}
\begin{equation}
\hspace{-0mm}
B'(S') = \eta\ P(Z|S') \sum_{S\in\set{S}}P(S'|S,A) B(S)
\label{eq:belup}
\vspace{-0mm}
\end{equation}
where $\eta\triangleq 1/P(Z|B,A)$ is a normalizing constant.
A policy $\pi$ for the POMDP controller is defined as a mapping from each belief $B$ to an action $A$ (Section~\ref{sec:polcom}). Solving a POMDP involves choosing the optimal policy $\pi^\ast$ that maximizes the expected reward for any given belief $B$: 
\vspace{-0mm}
\begin{equation*}
\hspace{-0mm}
\begin{array}{rl}
\displaystyle \pi^\ast (B) = \hspace{-0mm} \displaystyle\mathop{\arg\max}_{A\in\set{A}}\sum_{Z'\in\set{Z}}{R(B')P(Z'|B,A)}\ .
\end{array}
\label{eq:policy}
\vspace{-0mm}
\end{equation*}
When the number of targets and active cameras increases, the state space and hence the belief space of the POMDP grow exponentially (Section~\ref{sec:staact}). Therefore, computing the optimal policy incurs exponential time. Fortunately, by exploiting the structure of our surveillance problem (Sections~\ref{sec:tramod} and \ref{sec:obsmod}), 
the optimal policy for a given belief $B$ can be computed efficiently (Section~\ref{sec:polcom}).
\subsection{States, Actions, and Observations}
\label{sec:staact}
\vspace{-0mm}

A joint state $S\in\set{S}$ of the POMDP controller is defined as a pair of joint states $T\in\set{T}^m$ of $m$ targets and $C\in\set{C}^n$ of $n$ active cameras where $\set{T}$ and $\set{C}$ denote sets of all possible states of each target and active camera, respectively. 
That is, $S \triangleq (T,C)$ and $\set{S}=\set{T}^m \times \set{C}^n$.
Let $T \triangleq (t_1,t_2, \ldots,t_m) \in \set{T}^m$ and $C \triangleq (c_1,c_2,\ldots,c_n) \in \set{C}^n$ where $t_k \in \set{T}$ and $c_i \in \set{C}$ denote the corresponding states of target $k$ and camera $i$. 
Let $t_k \triangleq (t_{l_k}, t_{d_k}, t_{v_k})\in\set{T}_l\times \set{T}_d\times \set{T}_v$ where $t_{l_k}, t_{d_k}$, and $t_{v_k}$ denote target $k$'s location, direction, and velocity, respectively. That is, $\set{T} = \set{T}_l\times \set{T}_d\times \set{T}_v$.

The state space $\set{C}$ of an active camera is a finite set of discrete pan/tilt/zoom positions. 
Let $fov(c_i) \subset \set{T}_l$ be a subset of target locations lying within the fov of camera $i$ in its state $c_i$. The joint fov of all cameras in joint state $C$ is defined as $fov(C) = \bigcup_{i=1}^{n} fov(c_i)$. 
The depth of fov of each active camera is limited such that (a) imageries of the targets detected within its fov satisfy a pre-defined resolution, and (b) the observed locations of the targets detected within its fov are of minimal location error. This is done by adjusting the zoom parameter of each camera based on its position.



The joint actions of the POMDP controller are PTZ commands that move the corresponding cameras to their specified states.
Let a joint action of the $n$ cameras be denoted by $A\triangleq (a_1, a_2, \ldots , a_n) \in \set{A}$ where $a_i$ denotes the PTZ command of camera $i$.

Let $\dot{\set{Z}}\triangleq\set{T}_l \cup \{\phi\}$ denote a set of all possible observations of a target comprising the set $\set{T}_l$ of all possible locations of the target in the environment and a null observation $\phi$ when the target is not observed by any of the cameras. Let an observation of target $k$
be denoted by $z_k \in \dot{\set{Z}}$ and a joint observation of the $m$ targets be denoted by $Z\triangleq (z_1, z_2, \ldots, z_m) \in \dot{\set{Z}}^m$.
That is, $\set{Z} = \dot{\set{Z}}^m$.
\subsection{Transition Model $T_f$}
\label{sec:tramod}
\vspace{-0mm}
By exploiting the following structural assumptions in the state transition dynamics of the surveillance environment:
\squishlisttwo
\item camera $i$'s next state $c'_i$ is conditionally independent of the other $n-1$ cameras' states and actions and the $m$ targets' states given its current state $c_i$ and action $a_i$ for $i=1,\ldots, n$ and
\item target $k$'s next state $t'_k$ is conditionally independent of the $n$ cameras' states and actions (i.e., target's motion is not affected by the cameras' states and actions) and the other $m-1$ targets' states (i.e., every target moves independently) given its current state $t_k$ for $k=1,\ldots, m$,
\squishend
the transition model $T_f$ can be factored into transition models of individual targets and active cameras, hence significantly reducing the time incurred to compute the optimal policy $\pi^\ast$ for a given belief $B$ (Section~\ref{sec:polcom}). Furthermore, since the modern active cameras are able to move to their specified positions accurately \cite{Axis_2011}, it is practical to assume the transition model of each individual camera to be deterministic and consequently represented by a function 
$\tau$
that moves camera $i$ from its current state $c_i$ to its next state $\tau(c_i,a_i)$ by the action $a_i$. Then, the transition model of the POMDP controller can be simplified to
%
\vspace{-0mm}
\begin{equation}
P(S'|S, A) = \displaystyle \hspace{-0mm}\prod_{k = 1}^{m}P(t_{k}'|t_k) \prod_{i = 1}^{n} \delta_{\tau(c_i,a_i)}(c'_i)
\label{eq:4}
\vspace{-0mm}
\end{equation}
where $\delta_x(x')$ is a Kronecker delta function of value $1$ if $x' = x$, and $0$ otherwise.
Details on the derivation of (\ref{eq:4}) are reported in \cite{Natarajan_2012}.
The state transition of target $k$ from $t_k$ to $t'_k$ includes stochastic transitions of its location from $t_{l_k}$ to $t'_{l_k}$, its direction from $t_{d_k}$ to $t'_{d_k}$, and its velocity from $t_{v_k}$ to $t'_{v_k}$. So, the transition probability of target $k$ can be factored into transition probabilities of its location, direction, and velocity:
\vspace{-0mm}
\begin{equation*}
P(t_k'|t_k)
=\displaystyle P(t'_{l_k} | t_{l_k}, t'_{d_k}, t'_{v_k}) P(t'_{d_k} | t_{d_k}) P(t'_{v_k} | t_{v_k})\ .
\vspace{-0mm}
\label{eq:targettransmodel}
\end{equation*}
The transition probabilities $P(t'_{d_k} | t_{d_k})$ and $P(t'_{v_k} | t_{v_k})$ of the target's direction and velocity are, respectively, modeled as Gaussian distributions $\set{N}(\mu_d,\sigma_d)$ and $\set{N}(\mu_v,\sigma_v)$ with the means $\mu_d$ and $\mu_v$ being the current direction and velocity of the target, and $\sigma_d$ and $\sigma_v$ being the variance parameters which are learned from a dataset of the targets' trajectories in the environment. The transition probability $P(t'_{l_k} | t_{l_k}, t'_{d_k}, t'_{v_k})$ of the target's next location is constructed using the general velocity-direction motion model, as described in \cite{Natarajan_2012}. 
\subsection{Observation Model {$O_f$}}
\label{sec:obsmod}
\vspace{-0mm}
Similar to the factorization of the transition model $T_f$, 
the observation model $O_f$ can also be factored into observation models of individual targets using the following structural assumption: The observed location $z_k \in \dot{\set{Z}}$ of target $k$ is conditionally independent of the observed and true states of the other $m-1$ targets and its true direction $t_{d_k} \in \set{T}_d$ and velocity $t_{v_k} \in \set{T}_v$
given its true location $t_{l_k} \in \set{T}_l$ and the joint state $C\in\set{C}^n$ of the $n$ active cameras for $k =1,\ldots,m$.
As a result, the time incurred to compute the optimal policy $\pi^\ast$ for a given belief $B$ can be significantly reduced (Section~\ref{sec:polcom}). 
Then, the observation model of the POMDP controller can be simplified to
\begin{equation}
\displaystyle \hspace{-0mm} P(Z|S)\hspace{-0mm} 
         = \prod_{k=1}^{m}{P(z_k|t_{l_k},C)}\ .
\label{eq:obsmodel}
\end{equation}
%
\ifthenelse{\value{sol}=1}{The derivation of (\ref{eq:obsmodel}) is reported in Appendix~\ref{ap:proof1}.}
{The derivation of (\ref{eq:obsmodel}) is reported in \cite{Natarajan_2012_ext}.}
The observation probability $P(z_k|t_{l_k},C)$ of target $k$ depends on whether the target lies within the joint fov of the active cameras.
When the target lies within the cameras' joint fov corresponding to their joint state $C$ (i.e., $z_k\neq\phi$), the observation model of target $k$ becomes deterministic:
\begin{equation*}
P(z_k|t_{l_k},C)=\left\{ 
				\begin{array}{cl}
				\displaystyle\hspace{-0mm}1 &\hspace{-0mm} \mbox{if $z_k=t_{l_k} \wedge t_{l_k}\in fov(C),$}\\
				\hspace{-0mm} 0 & \hspace{-0mm} \mbox{otherwise.}
				\end{array}\right.
\label{eq:targetobsprob1}
\end{equation*}
On the other hand, when target $k$ does not lie within the joint fov of the active cameras corresponding to their joint state $C$, the observation probability of target $k$ is uniformly distributed over the locations not covered by the joint fov (i.e., $\overline{fov(C)}$):
\begin{equation*}
P(z_k=\phi |t_{l_k},C)=\left\{ 
				\begin{array}{cl}
				\displaystyle\hspace{-0mm}\frac{1}{|\overline{fov(C)}|} &\hspace{-0mm}\mbox{if $t_{l_k}\notin fov(C),$}\vspace{0mm}\\
				\displaystyle\hspace{-0mm}0 &\hspace{-0mm}\mbox{otherwise.}	
				\end{array}\right.
\label{eq:targetobsprob2}
\end{equation*}

\subsection{Bayesian Belief Update}
\label{sec:beltar}
By making use of independence assumptions similar to that in the transition model (Section~\ref{sec:tramod}), 
a belief $B$ can be factored into beliefs of individual targets and cameras:
\begin{equation}
\begin{array}{rl}
B(S) =& \hspace{-2mm}P((T, C)) = P(T)P(C)\\
=& \hspace{-2mm}\displaystyle\prod_{k = 1}^{m}P(t_k)\prod_{i = 1}^{n}P(c_i)
= \prod_{k = 1}^{m}b_k(t_k)\prod_{i = 1}^{n}\delta_{\hat{c}_{i}}(c_i) 
\end{array}
\label{eq:bel}
\end{equation}
where $b_k$ denotes a belief over the set $\set{T}$ of all possible states of target $k$ (i.e., $b_k(t_k)$ is the probability that target $k$ is in state $t_k$) and $\hat{c}_{i}$ is the current state of camera $i$ that,
unlike a target's state, is fully observable to the POMDP controller since its position can be directly read from its port.
Hence, the probability $P(c_i)$ of a state $c_i$ of camera $i$ can be represented by a
Kronecker delta $\delta_{\hat{c}_{i}}(c_i)$ and the last equality in (\ref{eq:bel}) follows.



The POMDP controller issues a joint action $A$ to move each camera $i$ from current state $\hat{c}_{i}$ to next state $\hat{c}'_{i}$, receives an observation $z_k$ of each target $k$,
 and then updates the prior belief $B$ to the posterior belief $B'$ using Bayes' rule (\ref{eq:belup}). 
Similar to the factorization of the prior belief $B$ above,
the posterior belief $B'$ can also be factored into posterior beliefs of individual targets and cameras:
\begin{equation}
B'(S') = \prod_{k=1}^{m}b'_{k}(t'_k) \prod_{i = 1}^{n}\delta_{\hat{c}'_{i}}(c'_i) 
\label{eq:belfact1}
\end{equation}
where the posterior belief $b'_{k}$ of target $k$ is defined as
\begin{equation}
b'_{k}(t'_k) 
\triangleq \eta_k P(z_k| t'_{l_k}, C') \sum_{t_k\in\set{T}}{P(t'_k|t_k)b_k(t_k)}\ ,
\label{eq:belup2}
\end{equation}
$C'\triangleq (c'_1,\ldots,c'_n)$, and 
${\eta_k}\triangleq 1/P(z_k|b_k,C')$
is a normalizing constant. 
\ifthenelse{\value{sol}=1}{The derivation of (\ref{eq:belfact1}) is reported in Appendix~\ref{ap:proof2}.}
{The derivation of (\ref{eq:belfact1}) is reported in \cite{Natarajan_2012_ext}.}



\subsection{Objective/Reward Function {$R$}}
\label{sec:objfun}
The goal of the surveillance system is to maximize the number of targets observed with a guaranteed resolution. This can be achieved by defining a reward function that measures the total number of targets lying within the joint fov of the active cameras corresponding to their joint state $C$:
\begin{equation}
R(S) = R((T, C)) \triangleq \displaystyle \sum_{k = 1}^{m}{\widetilde{R}(t_{k},C)} 
\label{eq:rew_max1}
\end{equation}
where
\begin{equation*}
\hspace{-0mm}
\displaystyle \widetilde{R}(t_{k},C) \triangleq \left\{ 
						\begin{array}{ll}
								1 & \mbox{if } t_{l_k}\in fov(C),\\ 
								0 & \mbox{otherwise.}\end{array} 
						\right.
\label{eq:rew_max2}
\end{equation*}
Since the exact locations of the targets may not be fully observable to the cameras at all times, the POMDP controller has to track the joint belief $B$ of the targets and consider the \emph{expected} reward with respect to this belief instead:
\begin{equation}
\displaystyle R(B)  \triangleq\sum_{S\in \set{S}}R(S)B(S)
=\sum_{k=1}^{m}\widetilde{R}(b_k,\widehat{C})
\label{eq:rew3}
\end{equation}
where $\widehat{C}\triangleq(\hat{c}_{1},\ldots,\hat{c}_{n})$ and
\begin{equation}
\displaystyle\widetilde{R}(b_k,C)\triangleq\sum_{t_k\in\set{T}}{\widetilde{R}(t_k,C)b_k(t_k)}\ .
\label{bry}
\end{equation}
\ifthenelse{\value{sol}=1}{The derivation of (\ref{eq:rew3}) is reported in Appendix~\ref{ap:proof3}.}
{The derivation of (\ref{eq:rew3}) is reported in \cite{Natarajan_2012_ext}.}
%
\section{Policy Computation}
\label{sec:polcom}
\begin{figure*}
\begin{tabular}[h]{cccc}
\hspace{0mm}\includegraphics[width=0.22\textwidth]{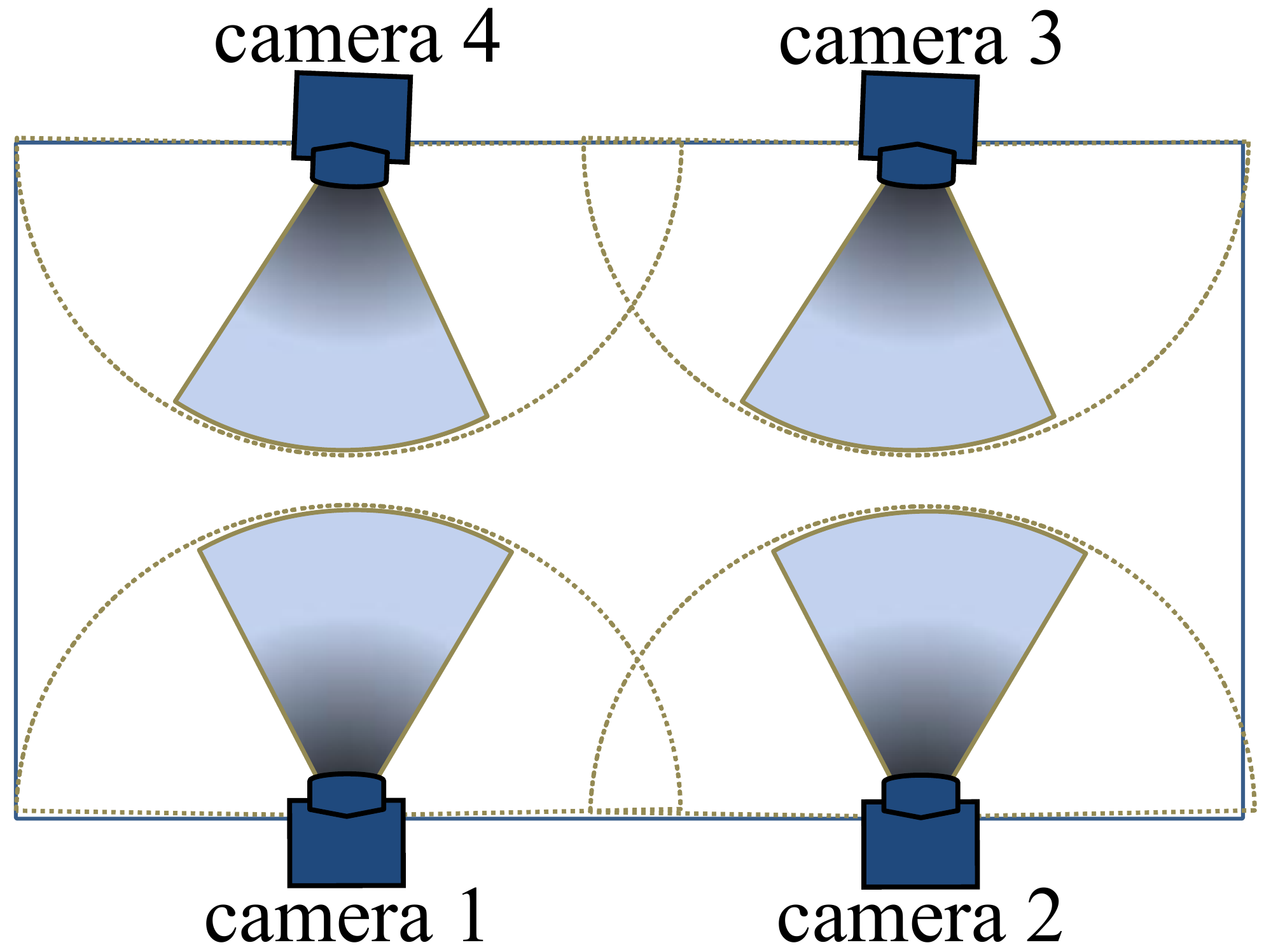} & \hspace{-0mm}\includegraphics[width=0.24\textwidth]{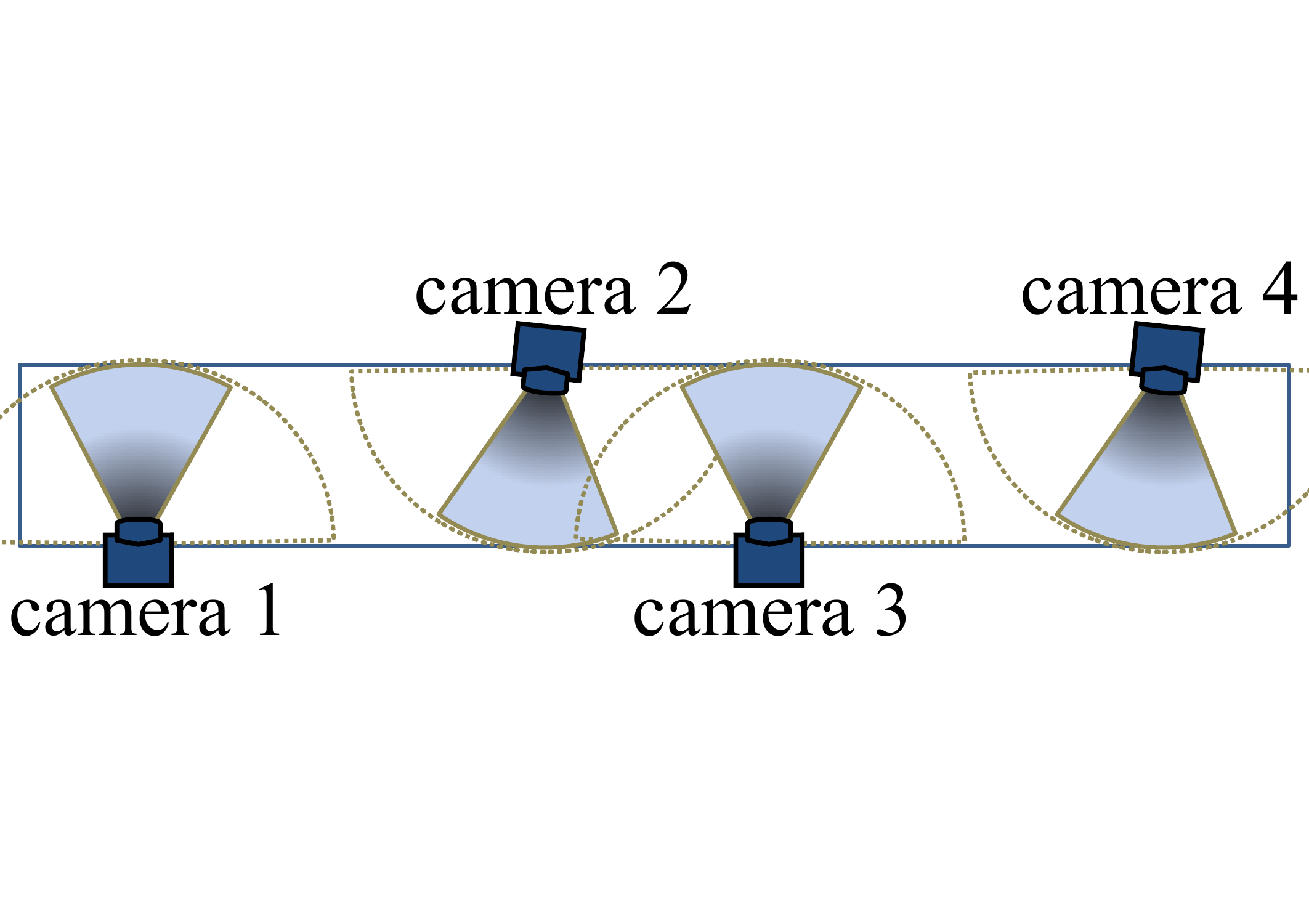} & \hspace{-0mm}\includegraphics[width=0.22\textwidth]{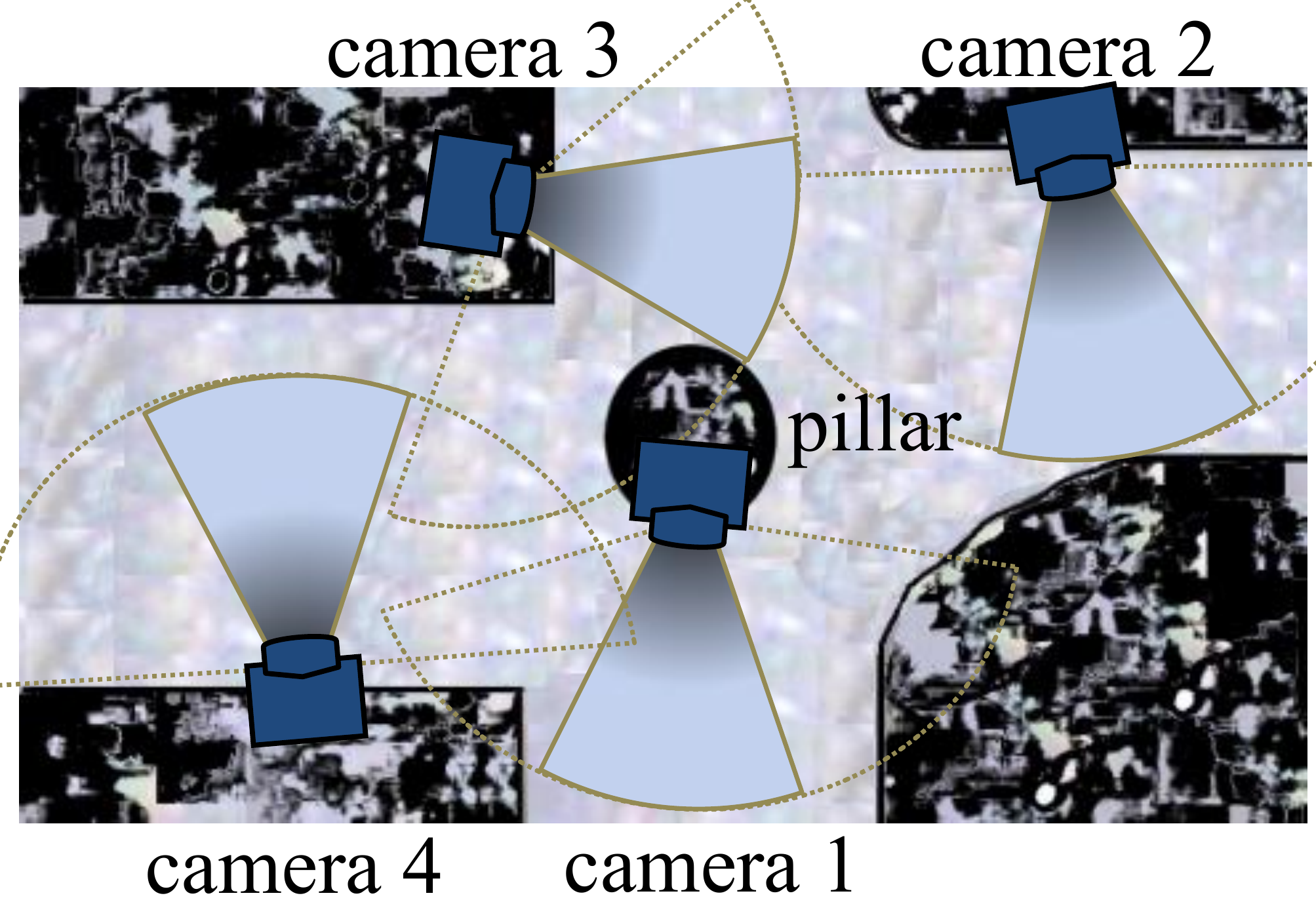} & \hspace{-0mm}\includegraphics[width=0.22\textwidth]{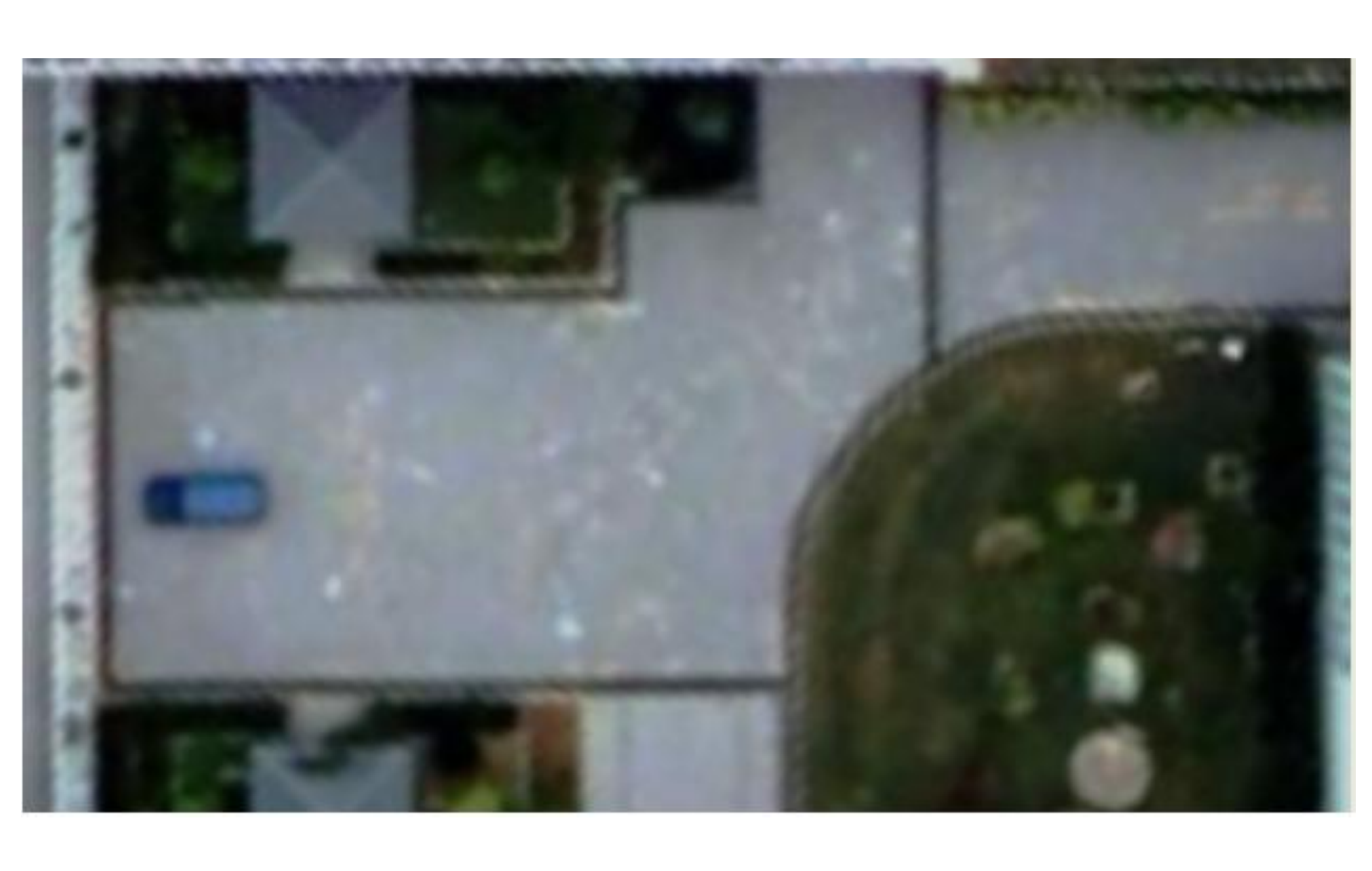}\vspace{-1mm}\\
(a)&(b)&(c)&(d)
\vspace{-1mm}
\end{tabular}
\caption{Setups of simulated surveillance environments: (a) Hall ($|\set{T}_l|=20 \times 10$ target locations), (b) corridor ($|\set{T}_l|=40 \times 5$ target locations), and (c) junction ($|\set{T}_l|=168$ target locations) with its corresponding real-world map shown in (d).}
\label{fig:expsetups}
\end{figure*}
Recall that a policy $\pi$ for the POMDP controller is a mapping from each belief $B$ to a joint action $A\in\set{A}$ of the $n$ cameras. At every time step, the POMDP controller determines the optimal policy $\pi^\ast$ for the belief $B$ such that the expected number of observed targets in the next time step is maximized. Since the observations of the $m$ targets taken by the cameras in the next time step are not known to the POMDP controller, it has to consider the \emph{expected} reward with respect to these future observations. Then, the optimal policy $\pi^\ast$ for a given belief $B$ becomes
\begin{equation}
\displaystyle \pi^\ast (B) = \hspace{-0mm} \displaystyle\mathop{\arg\max}_{A\in\set{A}} V(B,A)
\label{eq:policy1}
\end{equation}
where 
\begin{equation}
\displaystyle V(B,A) = \displaystyle\sum_{Z\in\set{Z}}{R(B')P(Z|B,A)}\ .
\label{eq:policy2}
\end{equation}
Computing the policy $\pi^\ast$ (\ref{eq:policy1}) for a given belief $B$ incurs $\set{O}(|\set{A}||\dot{\set{Z}}|^m|\set{T}|)$ time which is exponential in the number $m$ of targets. 
Fortunately, by exploiting simplified transition and observation models due to conditional independence assumptions (i.e., (\ref{eq:4}) and (\ref{eq:obsmodel})), this computational cost can be significantly reduced. In particular,  
\ifthenelse{\value{sol}=1}{it is derived in Appendix~\ref{ap:proof4}}{it can be derived\cite{Natarajan_2012_ext}}
that the value function $V(B, A)$ of $m$ targets can be simplified to comprise a sum of value function $\widetilde{V}(b_k, C')$ of individual target $k$ for $k=1,\ldots,m$:
\begin{equation}
 V(B, A) = \displaystyle\sum_{k=1}^{m}{\widetilde{V}(b_k,(\tau(\hat{c}_{1},a_1),\ldots,\tau(\hat{c}_{n},a_n)))}
\label{eq:policy3}
\end{equation}
where
\begin{equation}
\widetilde{V}(b_k,C')\triangleq\displaystyle\sum_{z_k\in  fov(C')}\sum_{t'_k\in\set{T}}\widetilde{R}(t'_k,C')\hat{b}'_k(t'_k)
\label{eq:policy4}
\end{equation}
and $\hat{b}'_k$ is the unnormalized belief of target $k$ (i.e., $\hat{b}'_k(t'_k)={b}'_k(t'_k)/\eta_k$). 
Using (\ref{eq:policy3}) and (\ref{eq:policy4}), we obtain the following result:\vspace{1mm}
 \begin{thm}
\emph{If (\ref{eq:4}) and (\ref{eq:obsmodel})
hold, then computing policy $\pi^\ast$ (\ref{eq:policy1}) for a given belief $B$ incurs $\set{O}( |\set{A}||\dot{\set{Z}}||\set{T}|m)$ time.}\vspace{1mm}
\label{thm1}
\end{thm}
Computing the value function $\widetilde{V}(b_k,C')$ (\ref{eq:policy4}) for a single target $k$ incurs $\set{O}(|\dot{\set{Z}}||\set{T}|)$ time. For $m$ targets, the value function $V(B,A)$ (\ref{eq:policy3}) therefore incurs $\set{O}(|\dot{\set{Z}}||\set{T}|m)$ time. Finally, computing the optimal policy $\pi^\ast$ (\ref{eq:policy1}) for a given belief $B$ incurs $\set{O}( |\set{A}||\dot{\set{Z}}||\set{T}|m)$ time which is linear in number $m$ of targets.
%
\section{Experiments and Discussion}
\label{expdis}
This section evaluates the performance of our proposed POMDP controller in simulations over various realistic surveillance environments using Player/Stage simulator \cite{Gerkey_2003} and with real AXIS 214 PTZ cameras to show its practical feasibility in real-world surveillance. 
Our POMDP-based approach (denoted by $P$ in Fig.~\ref{fig:graphsmallfov}) that uses only active PTZ cameras is compared against the following state-of-the-art multi-camera coordination and control techniques under partially observable surveillance environments:
\squishlisttwo
\item \emph{MDP with only PTZ cameras ($MP$)}:
This approach uses a Markov Decision Process (MDP) controller \cite{Natarajan_2012} to coordinate and control the active cameras.  
There is no static camera to directly observe the targets' locations. Hence, they are observed only from the active cameras' fov; 
\item \emph{MDP with static and PTZ cameras ($MSP$)}:
This approach uses the MDP controller \cite{Natarajan_2012} to coordinate and control the active cameras that are supported by wide-view static cameras. 
A Gaussian noise is added to the location of each target observed by the static camera
such that the Gaussian variance increases with greater distance of the target from the static camera; 
\item \emph{Systematic Approach ($Sys$)}:
The active cameras are panned systematically to each of its states in a round robin fashion for every time step; and
\item \emph{Static Approach ($Stat$)}:
The active cameras are fixed at a particular state such that they observe the maximum area of the environment.
\squishend
The performance metric used to evaluate the above approaches is given by
\begin{equation*}
\displaystyle PercentObs = \frac{100}{\tau M_{tot}}\sum^{\tau}_{i=1} {M^{i}_{obs}} 
\end{equation*}
where $\tau$ (i.e., set to $100$ in simulations) is the total number of time steps taken in our experiments, 
$M^{i}_{obs}$ is the total number of targets observed by the active cameras at a given time step $i$, 
and $M_{tot}$ is the total number of targets present in the environment. That is, the $PercentObs$ metric averages the percentage of targets being observed by the active cameras over the entire duration of $\tau$ time steps. 
%
\subsection{Simulated Experiments}
\label{sec:simexp}
Fig.~\ref{fig:expsetups} shows three different setups of simulated surveillance environments: (a) hall, (b) corridor, and (c) junction. The junction setup simulates a surveillance environment within our university campus which consists of obstacles (black shades in Fig.~\ref{fig:expsetups}c) like buildings and walls. In order to introduce more occlusions in the environment, we have added a virtual pillar in the center of the junction setup (Fig.~\ref{fig:expsetups}c). 
The active cameras are simulated in Player/Stage simulator by configuring the states of the cameras across various pan angles, as discussed in Section~\ref{sec:staact}. There are $n=4$ active cameras with $|\set{C}|=3$ states each. 
The targets' trajectories are generated in the simulator based on the velocity-direction motion model (Section~\ref{sec:tramod}) which resembles real human motion in a surveillance environment. Every target can move in one of the $8$ possible discretized directions $\set{T}_d=\{0\,^{\circ}, 45\,^{\circ}, \ldots,-90\,^{\circ},-45\,^{\circ}\}$ with an assumed velocity of $1.5$ cells per time step. 
The transition model, observation model, and reward function for a single target are computed and stored offline for the above setups. 

Fig.~\ref{fig:graphsmallfov} shows the comparison of performance of different approaches for up to $m=20$ targets for all three setups. It can be observed that our POMDP controller outperforms the other evaluated approaches in all three setups. The detailed observations from the experiments are as follows:
\begin{figure}
\begin{tabular}{ccc}
\hspace{-2mm}\includegraphics[height=0.125\textwidth]{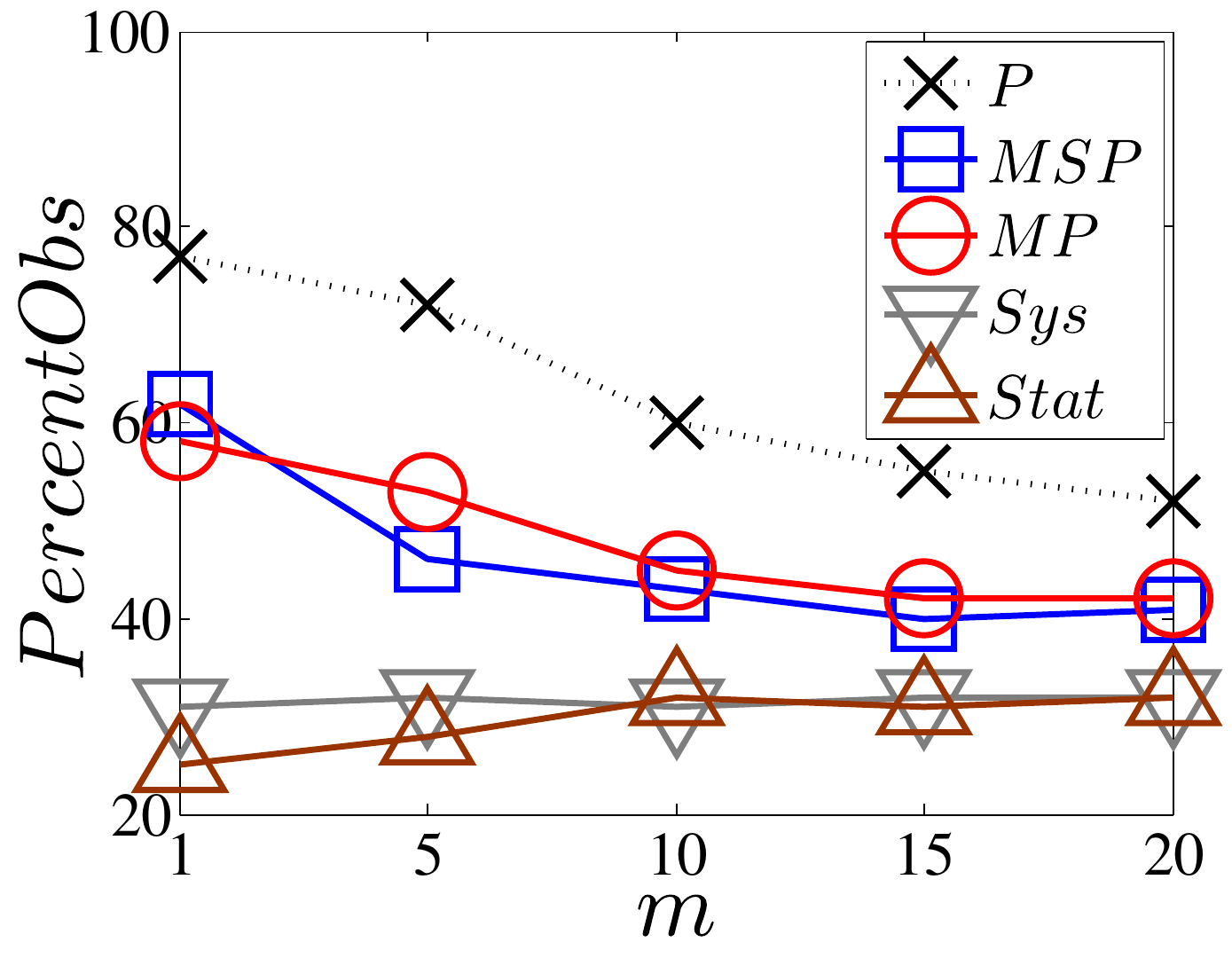} & \hspace{-3mm}\includegraphics[height=0.125\textwidth]{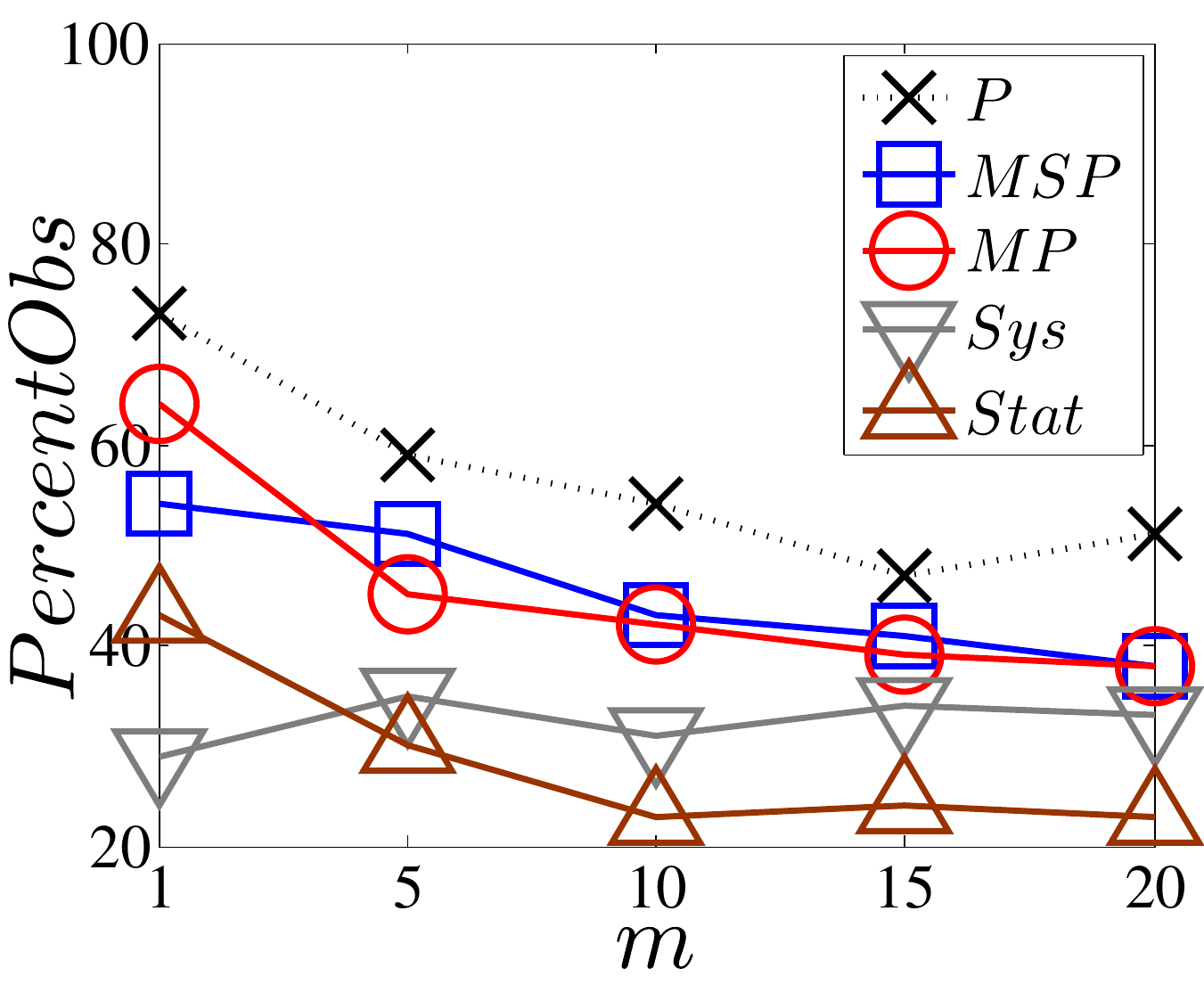} & \hspace{-3mm}\includegraphics[height=0.125\textwidth]{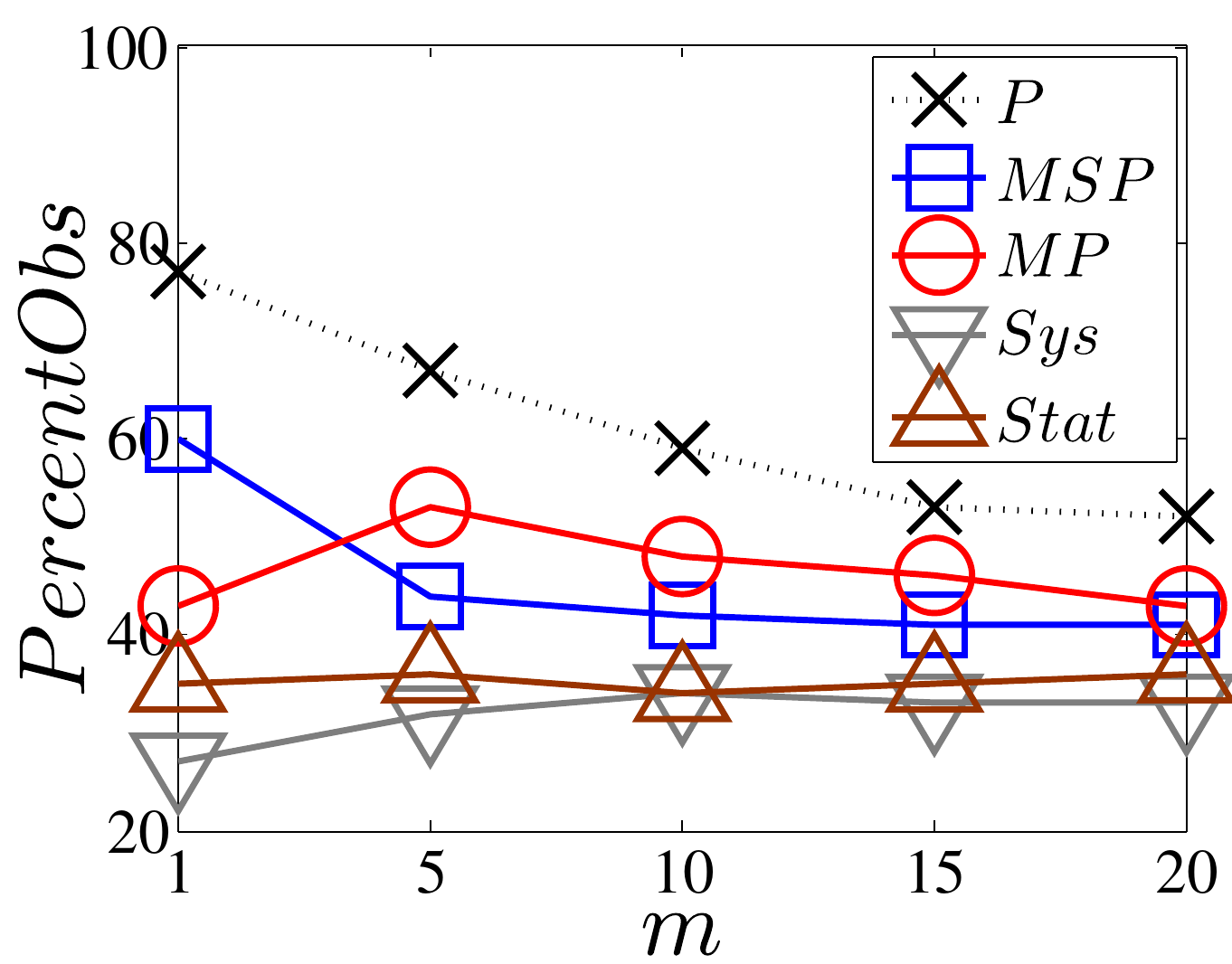} \vspace{-1mm}\\
\hspace{1mm}{(a)}&\hspace{-0.5mm}{(b)}&\hspace{0mm}{(c)}
\vspace{-1mm}
\end{tabular}
\caption{Graphs of $PercentObs$ vs. number $m$ of targets with $n=4$ active cameras for the (a) corridor, (b) hall and (c) junction setups.}
\label{fig:graphsmallfov}
\end{figure}

Our POMDP controller performs better than the $MP$ approach because (a) when the targets leave the fov of any of the cameras and enter an occluded region, the active cameras in the $MP$ approach have no idea where the targets will be moving to in the next few time steps, and (b) when the targets enter the fov of any of the active cameras from an occluded region, the directions of the targets are wrongly interpreted by the MDP controller. This is a serious limitation of the $MP$ approach, i.e., there is no way of knowing the direction of the targets when they are in an occluded region. In contrast, the Bayesian belief update process in our POMDP controller helps to trace the locations and directions of the targets, even when they are not observed in any of the cameras. Hence, the cameras are controlled based on the belief of the targets. 

Our POMDP controller performs better than the $MSP$ approach because when the static cameras observe the targets that are far away, they obtain noisy locations of the targets. This in turn induces the errors in the direction and velocity of the targets. 
Hence, when the  noisy targets' information is used in the MDP controller, it predicts the expected locations of the targets poorly, which consequently affects the performance of the $MSP$ approach. In contrast, for our POMDP-based approach, the targets' locations are observed by high-resolution active cameras whose calibration error is bounded by limiting the depth of its fov (Section~\ref{sec:staact}). Since the observations (i.e., locations of the targets) for POMDP are more accurate than in the $MSP$ approach, the predictions of the targets' locations and directions through the Bayesian belief update process are also more accurate. 

Our POMDP controller performs much better than the $Sys$ and $Stat$ baseline approaches because, for our approach, the active cameras are controlled based on the targets' predicted motion and observations taken by the active cameras. But, for the $Sys$ approach, the cameras are panned without accounting for the targets' information such as locations and direction while, for the $Stat$ approach, every camera is fixed in one of the states. 

To summarize, 
our POMDP-based approach performs better than the $MP$ approach due to its ability to keep track of the targets' locations and directions through its Bayesian belief update process.
It outperforms the $MSP$ approach because the observations (i.e., target's location) taken by the active cameras in our POMDP controller are more accurate as compared to the noisy observations taken by the static cameras in the $MSP$ approach. 
Lastly, the $Sys$ and $Stat$ approaches suffer from performance degradation because the cameras are controlled independently of the targets' information.

\subsection{Real Camera Experiments}
\label{sec:realexp}
\begin{table}[tbp]
	\centering		
	\caption{Performance for real camera experiments.}
	\vspace{-3mm}
				\begin{tabular}{|c|c|c|c|c|c|}
				\hline
				No. of targets $m$ & 1 & 2 & 3 & 4 & 5\\
				\hline
				$PercentObs$ & 98.2 & 96.6 & 93.3 & 91.5 & 87\\
				\hline
			\end{tabular}
	\label{tab:realexp}
\end{table}
The feasibility of our POMDP controller is tested using real AXIS $214$ PTZ cameras to monitor Lego robots (targets) over the environment of size $|\set{T}_l|=10\times8$ grid cells. We have $n=3$ PTZ cameras, each of which has $|\set{C}|=3$ states. These cameras are calibrated in each of its state and the depth of the fov of these  cameras are determined empirically for each of them. The Lego robots are programmed to move based on the velocity-direction motion model. Table~\ref{tab:realexp} shows the performance of our approach in real camera experiments. Due to space limitation, we showcase our detailed results of real camera experiments in a demo  video\footnote{{http://www.comp.nus.edu.sg/$\sim$lowkh/camera.html}\vspace{-5mm}}. 
%

The limitations of our POMDP-based approach are as follows: (a) it scales well only in number of targets and needs improvement in scalability in the number of cameras; and (b) it works well only if the underlying computer vision algorithms for target detection and recognition perform accurately. For our future work, we would like to extend this work by scaling to a large number of cameras and also accounting for the uncertainties arising from the underlying vision algorithms. We would also like to deploy active cameras along with a team of robots (\cite{LowAAAI04,LowSMCC06}) for indoor surveillance.
\section{Conclusion}
This paper describes a novel POMDP-based approach to coordinating and controlling a network of active cameras for maximizing the number of targets observed with a guaranteed resolution in an uncertain, partially observable surveillance environment.
Specifically, our approach helps to eliminate the dependency on wide-view static cameras for tracking the targets' locations and simultaneously performs the tracking and observation of the targets at high resolution. We have exploited the conditional independence property in the targets' motion and observation for our surveillance problem in order to reduce the exponential policy computation time to linear time in the  number of targets. The experimental evaluation shows that our proposed POMDP controller performs better than the state-of-the-art approaches and is feasible and practical to operate in real-world environments. 
\bibliographystyle{abbrv}
\bibliography{sigproc} 
\ifthenelse{\value{sol}=1}{
\clearpage
\appendix
\subsection{Observation model factorization}
\label{ap:proof1}
\begin{equation*}
\begin{array}{l}
\displaystyle  P(Z|S)\\
= \displaystyle P(Z|T,C)\\ 
= \displaystyle P(z_1,z_2,\ldots,z_m|t_1,t_2,\ldots,t_m,C)\\
= \displaystyle \prod_{k=1}^{m}{P(z_k|t_{k},C)}\\
=\displaystyle \prod_{k=1}^{m}{P(z_k|t_{l_k},C)}\ .
\end{array}         
\label{eq:obsmodel_ap}
\end{equation*}
The last two equalities are due to the conditional independence assumption in the observation model (Section~\ref{sec:obsmod}).

\subsection{Posterior belief decomposition}
\label{ap:proof2}
\begin{equation*}
\begin{array}{l}
B'(S')\\
\displaystyle= \eta\ P(Z|S') \sum_{S\in\set{S}}P(S'|S,A) B(S)\vspace{1mm}\\
\displaystyle= \eta\ \prod_{k=1}^{m}{P(z_k|t'_{l_k},C')} \sum_{T\in\set{T}^m}\sum_{C\in\set{C}^n}\prod_{k = 1}^{m}P(t_{k}'|t_k)\vspace{1mm}\\
\displaystyle\hspace{7.7mm}\prod_{i = 1}^{n} \delta_{\tau(c_i,a_i)}(c'_i)
\prod_{k = 1}^{m}b_k(t_k)\prod_{i = 1}^{n}\delta_{\hat{c}_{i}}(c_i) \vspace{1mm}\\
\displaystyle= \eta\ \prod_{k=1}^{m}{P(z_k|t'_{l_k},C')} \left(\sum_{T\in\set{T}^m}\prod_{k = 1}^{m}P(t_{k}'|t_k)\prod_{k = 1}^{m}b_k(t_k)\right)\vspace{1mm}\\
\displaystyle\hspace{7.0mm}\left(\sum_{C\in\set{C}^n}\prod_{i = 1}^{n} \delta_{\tau(c_i,a_i)}(c'_i)\prod_{i = 1}^{n}\delta_{\hat{c}_{i}}(c_i)\right) \vspace{1mm}\\
\displaystyle= \eta\ \prod_{k=1}^{m}{P(z_k|t'_{l_k},C')} \prod_{k = 1}^{m}\sum_{t_k\in\set{T}} P(t_{k}'|t_k)b_k(t_k)\vspace{1mm}\\
\displaystyle\hspace{7.7mm}\prod_{i = 1}^{n} \delta_{\tau(\hat{c}_{i},a_i)}(c'_i) \vspace{1mm}\\
\displaystyle= \sum_{C'\in\set{C}^n}\prod_{k=1}^{m}{\eta_k}\prod_{i=1}^{n}{\delta_{\hat{c}_i'}(c'_i)}\vspace{1mm}\\
\displaystyle\hspace{5.0mm}\prod_{k=1}^{m} P(z_k| t'_{l_k}, C') \sum_{t_k\in\set{T}}{P(t'_k|t_k)b_k(t_k)} \prod_{i = 1}^{n}\delta_{\hat{c}'_{i}}(c'_i) \vspace{1mm}\\
\displaystyle= \prod_{k=1}^{m}\eta_k P(z_k| t'_{l_k}, C') \sum_{t_k\in\set{T}}{P(t'_k|t_k)b_k(t_k)} \prod_{i = 1}^{n}\delta_{\hat{c}'_{i}}(c'_i) \vspace{1mm}\\
\displaystyle= \prod_{k=1}^{m}{b'_{k}(t'_k) \prod_{i = 1}^{n}\delta_{\hat{c}'_{i}}(c'_i) }\ .
\end{array}
\label{eq:belfact}
\end{equation*}
The first equality is due to (\ref{eq:belup}).
The second equality follows from (\ref{eq:4}), (\ref{eq:obsmodel}), and
(\ref{eq:bel}).
The fifth equality follows from $\displaystyle\eta= \sum_{C'\in\set{C}^n}\prod_{k=1}^{m}{\eta_k}\prod_{i=1}^{n}{\delta_{\hat{c}_i'}(c'_i)}$  (Section~\ref{eqp}).
The last equality is due to (\ref{eq:belup2}).
\subsection{Reward function decomposition}
\label{ap:proof3}
\begin{equation*}
\begin{array}{l}
R(B)\vspace{1mm}\\
=\displaystyle \sum_{S\in \set{S}}R(S)B(S)\vspace{1mm}\\
=\displaystyle \sum_{(T,C)\in\set{S}}{R((T,C))B((T,C))}\vspace{1mm}\\
=\displaystyle \sum_{C\in\set{C}^n}\sum_{T\in\set{T}^m}\sum_{k=1}^{m}\widetilde{R}(t_k,C)\prod_{k = 1}^{m}b_k(t_k)\prod_{i = 1}^{n}\delta_{\hat{c}_{i}}(c_i)\vspace{1mm}\\
=\displaystyle \sum_{k=1}^{m}\sum_{t_k\in\set{T}}\widetilde{R}(t_k,\widehat{C})b_k(t_k)\hspace{-1mm}\sum_{T_{-k}\in\set{T}^{m-1}}\prod_{j\not=k}b_j(t_j)\vspace{1mm}\\
=\displaystyle \sum_{k=1}^{m}\sum_{t_k\in\set{T}}\widetilde{R}(t_k,\widehat{C})b_k(t_k)\vspace{1mm}\\
=\displaystyle\sum_{k=1}^{m}\left(\sum_{t_k\in\set{T}}\widetilde{R}(t_k,\widehat{C})b_k(t_k)\right)\vspace{1mm}\\
=\displaystyle \sum_{k=1}^{m}\widetilde{R}(b_k,\widehat{C})
\end{array}
\end{equation*}
where $T_{-k} = (t_1,\ldots, t_{k-1},t_{k+1},\ldots,t_m)$.
The third equality is due to (\ref{eq:bel}) and (\ref{eq:rew_max1}).
The fifth equality follows from our independence assumption similar to that in (\ref{eq:bel}) and the law of total probability:
$$\sum_{T_{-k}\in\set{T}^{m-1}}\prod_{j\not=k}b_j(t_j)=\sum_{T_{-k}\in\set{T}^{m-1}}P(T_{-k})=1\ .$$ 
\subsection{Value function decomposition}
\label{ap:proof4}
\begin{equation*}
\begin{array}{l}
\displaystyle V(B,A)\vspace{1mm}\\ 
= \displaystyle\sum_{Z\in\set{Z}}{R(B')P(Z|B,A)}\vspace{1mm}\\
=\displaystyle\sum_{Z\in\set{Z}}\sum_{k=1}^{m}{\widetilde{R}(b'_k,\widehat{C}')\prod_{j=1}^{m}P(z_j|b_j,\widehat{C}')}\vspace{1mm}\\
=\displaystyle\sum_{k=1}^{m}\sum_{Z\in\set{Z}}\widetilde{R}(b'_k,\widehat{C}')\prod_{j=1}^{m}P(z_j|b_j,\widehat{C}')\vspace{1mm}\\
=\displaystyle
\sum_{k=1}^{m}\sum_{z_k\in\dot{\set{Z}}}\widetilde{R}(b'_k,\widehat{C}') P(z_k|b_k,\widehat{C}')\hspace{-1mm} \sum_{Z_{-k}\in\dot{\set{Z}}^{m-1}}{\prod_{j\not= k}P(z_j|b_j,\widehat{C}')}\vspace{1mm}\\
=\displaystyle\sum_{k=1}^{m}\sum_{z_k\in  fov(\widehat{C}')} \widetilde{R}(b'_k,\widehat{C}') P(z_k|b_k,\widehat{C}')\vspace{1mm}\\
= \displaystyle\sum_{k=1}^{m}\sum_{z_k\in  fov(\widehat{C}')}\sum_{t'_k\in\set{T}}\widetilde{R}(t'_k,\widehat{C}')b'_k(t'_k)P(z_k|b_k,\widehat{C}')\vspace{1mm}\\
= \displaystyle\sum_{k=1}^{m}\sum_{z_k\in  fov(\widehat{C}')}\sum_{t'_k\in\set{T}}\widetilde{R}(t'_k,\widehat{C}')\hat{b}'_k(t'_k)\vspace{1mm}\\
=\displaystyle\sum_{k=1}^{m}{\widetilde{V}(b_k,\widehat{C}')}
=\displaystyle\sum_{k=1}^{m}{\widetilde{V}(b_k,(\tau(\hat{c}_{1},a_1),\ldots,\tau(\hat{c}_{n},a_n)))}
\end{array}
\label{eq:valdecomp}
\end{equation*}
where $\widehat{C}'\triangleq(\hat{c}'_{1},\ldots,\hat{c}'_{n})$ and $Z_{-k} = (z_1,\ldots, z_{k-1},z_{k+1},\ldots,z_m)$.
The first equality is due to (\ref{eq:policy2}).
The second equality is obtained using (\ref{eq:rew3}) and $\displaystyle \eta^{-1}= \sum_{C'\in\set{C}^n}\prod_{k=1}^{m}{\eta^{-1}_k}\prod_{i=1}^{n}{\delta_{\hat{c}_i'}(c'_i)}$ (Section~\ref{eqp}).
The fifth equality follows from $\displaystyle P(Z_{-k}|B_{-k},A)=\prod_{j\neq k} P(z_j|b_j,\widehat{C}')$ where $\displaystyle B_{-k}(S) = \prod_{j \neq k} b_j(t_j)\prod_{i = 1}^{n}\delta_{\hat{c}_{i}}(c_i)$ and then the law of total probability:
$\displaystyle\sum_{Z_{-k}\in\dot{\set{Z}}^{m-1}}{\prod_{j\not= k}\hspace{-0mm} P(z_j|b_j,\widehat{C}')}=1$.
Also, note that when $z_k \notin fov(\widehat{C}')$, $\widetilde{R}(b'_k,\widehat{C}')=0$.
 The sixth equality is due to (\ref{bry}). Since the normalizing constant of $b'_k(t'_k)$ is $1/P(z_k|b_k,\widehat{C}')$, the seventh equality follows.
 %
\subsection{Derivation of $\displaystyle \eta= \sum_{C'\in\set{C}^n}\prod_{k=1}^{m}{\eta_k}\prod_{i=1}^{n}{\delta_{\hat{c}_i'}(c'_i)}$}
\label{eqp}
\begin{equation*}
\begin{array}{l}
\eta^{-1}\displaystyle= P(Z|B,A)\vspace{1mm}\\
\displaystyle= \sum_{S'\in{S}}{P(Z|S')P(S'|B,A)}\vspace{1mm}\\
\displaystyle=\sum_{S'\in{S}}{P(Z|S')\sum_{S\in\set{S}}{P(S'|S,A)P(S|B)}}\vspace{1mm}\\
\displaystyle= \sum_{S'\in\set{S}}{P(Z|S') \sum_{S\in\set{S}}{P(S'|S,A) B(S)}}\vspace{1mm}\\
\displaystyle=\sum_{C'\in\set{C}^n}{\sum_{T'\in\set{T}^m}{\prod_{k=1}^{m}{P(z_k|t_{l_k}',C')}}}\vspace{1mm}\\
\displaystyle \hspace{4.2mm}\sum_{C\in\set{C}^n}{\sum_{T\in\set{T}^m}{{\left(\prod_{k=1}^{m}{P(t_k'|t_k)}\prod_{i=1}^{n}{\delta_{\tau(c_i,a_i)}(c'_i)}\right) \left(\prod_{k=1}^{m}{b_k(t_k)}\prod_{i=1}^{n}{\delta_{\hat{c}_i(c_i)}}\right)}}}\vspace{1mm}\\
\displaystyle= \sum_{C'\in\set{C}^n}{\sum_{T'\in\set{T}^m}{\prod_{k=1}^{m}{P(z_k|t_{l_k}',C')}\left(\sum_{T\in\set{T}^m}{\prod_{k=1}^{m}{P(t_k'|t_k)}\prod_{k=1}^{m}{b_k(t_k)}}\right)}}\vspace{1mm}\\
\displaystyle \hspace{4.7mm}
\left(\sum_{C\in\set{C}^n}{\prod_{i=1}^{n}{\delta_{\tau(c_i,a_i)}(c'_i)}\prod_{i=1}^{n}{\delta_{\hat{c}_i}(c_i)}}\right)\vspace{1mm}\\
\displaystyle=\sum_{C'\in\set{C}^n}{\sum_{T'\in\set{T}^m}{\prod_{k=1}^{m}{P(z_k|t_{l_k}',C')}\prod_{k=1}^{m}{\sum_{t_k\in\set{T}}{P(t_k'|t_k)b_k(t_k)}}}}\vspace{1mm}\\
\displaystyle \hspace{5.3mm}
	\prod_{i=1}^{n}{\delta_{\tau(\hat{c}_i,a_i)}(c'_i)}\vspace{1mm}\\
\displaystyle= \sum_{C'\in\set{C}^n}{\sum_{T'\in\set{T}^m}{\prod_{k=1}^{m}{P(z_k|t_{l_k}',C')}\prod_{k=1}^{m}{\sum_{t_k\in\set{T}}{P(t_k'|t_k)b_k(t_k)}}}}\vspace{1mm}\\
\displaystyle \hspace{5.3mm}
\prod_{i=1}^{n}{\delta_{\hat{c}_i'}(c'_i)}\vspace{1mm}\\
\displaystyle= {\sum_{C'\in\set{C}^n}\prod_{k=1}^{m}{\sum_{t_k'\in\set{T}}{P(z_k|t_{l_k}',C')}\sum_{t_k\in\set{T}}{P(t_k'|t_k)b_k(t_k)}}\prod_{i=1}^{n}{\delta_{\hat{c}_i'}(c'_i)}}\vspace{1mm}\\
\displaystyle= {\sum_{C'\in\set{C}^n}\prod_{k=1}^{m}{\eta_k^{-1}}\prod_{i=1}^{n}{\delta_{\hat{c}_i'}(c'_i)}}
\end{array}
\label{eq:etaproof}
\end{equation*}
where $\displaystyle\eta_k^{-1}={\sum_{t'_k\in\set{T}}{P(z_k|t_{l_k}',C') \sum_{t_k\in\set{T}}{P(t_k'|t_k)b_k(t_k)}}}=P(z_k|b_k,C')$. The fourth equality follows from (\ref{eq:4}), (\ref{eq:obsmodel}), and
(\ref{eq:bel}). It follows that
$\displaystyle\eta={\sum_{C'\in\set{C}^n}\prod_{k=1}^{m}{\eta_k}\prod_{i=1}^{n}{\delta_{\hat{c}_i'}(c'_i)}}$.
}{}
\end{document}